\documentclass{Interspeech}

\interspeechcameraready

\title{Question Generation for Assessing Early Literacy Reading Comprehension}
\author[affiliation={1}]{Xiaocheng}{Yang}
\author[affiliation={1}]{Sumuk}{Shashidhar}
\author[affiliation={1}]{Dilek}{Hakkani-Tür}

\affiliation{}{Univerisity of Illinois Urbana-Champaign}{USA}
\email{xy61@illinois.edu, sumuks2@illinois.edu, dilek@illinois.edu}

\usepackage{comment}

\begin{document}

\maketitle

\begin{abstract}
Assessment of reading comprehension through content-based interactions plays an important role in the reading acquisition process.
In this paper, we propose a novel approach for generating comprehension questions geared to K-2 English learners. Our method ensures complete coverage of the underlying material and adaptation to the learner's specific proficiencies, and can generate a large diversity of question types at various difficulty levels to ensure a thorough evaluation. We evaluate the performance of various language models in this framework using the FairytaleQA dataset as the source material.
Eventually, the proposed approach has the potential to become an important part of autonomous AI-driven English instructors.
\end{abstract}

% Related Works
% 1. Question Generation for Reading Comprehension Assessment by Modeling How and What to Ask
% https://aclanthology.org/2022.findings-acl.168/
% 2. Question Generation for English Reading Comprehension Exercises using Transformers
% https://iaiai.org/letters/index.php/liir/article/view/183
% 3. Question Generation for Reading Comprehension of Language Learning Test : -A Method using Seq2Seq Approach with Transformer Model
% https://ieeexplore.ieee.org/document/8959903
% 4. Reading Comprehension Question Creator
% https://www.cathoven.com/reading-comprehension-question-generator/
% 5. Automatic generation of short-answer questions in reading comprehension using NLP and KNN
% https://pmc.ncbi.nlm.nih.gov/articles/PMC10091335/

\section{Introduction}
\label{sec:introduction}
Engaging children in conversation during storybook reading by asking them to complete sentences or asking questions to help them understand the content, known as dialogic reading, is a powerful strategy to foster language development. Dialogic reading involves specific prompts to encourage the child to become the storyteller while the adult can act as a supportive listener and questioner. Xu et al.~\cite{Xu_Aubele_Vigil_Bustamante_Kim_Warschauer_2021} have shown that a conversational agent can mimic the advantages of dialogic reading with a human by increasing children's vocalizations related to the story, decreasing unrelated vocalizations, and boosting story understanding.

In this paper, we present a question generation approach that supports quick and easy comprehension question generation given any learning materials and adapts the difficulty level to kindergarten to second-grade students. It can be utilized by conversational agents for tutoring style applications to enable a dialogic reading experience.

% Evaluation of reading comprehension through content-related interactions is an essential component of reading acquisition process. Such assessment enables educators to ascertain the degree to which students are grasping the material, facilitating the customization of instructional approaches and the provision of targeted assistance to learners experiencing difficulties. The appraisal of 
% comprehension extends beyond mere word identification, encompassing the extraction of textual meaning, the integration of prior knowledge, and the drawing of logical inferences.

We evaluate our approach on the FariytaleQA dataset and show state-of-the-art results in terms of question generation methods.

\section{System}
\label{sec:system}

\begin{figure}[t]
  \centering
  \includegraphics[width=\linewidth]{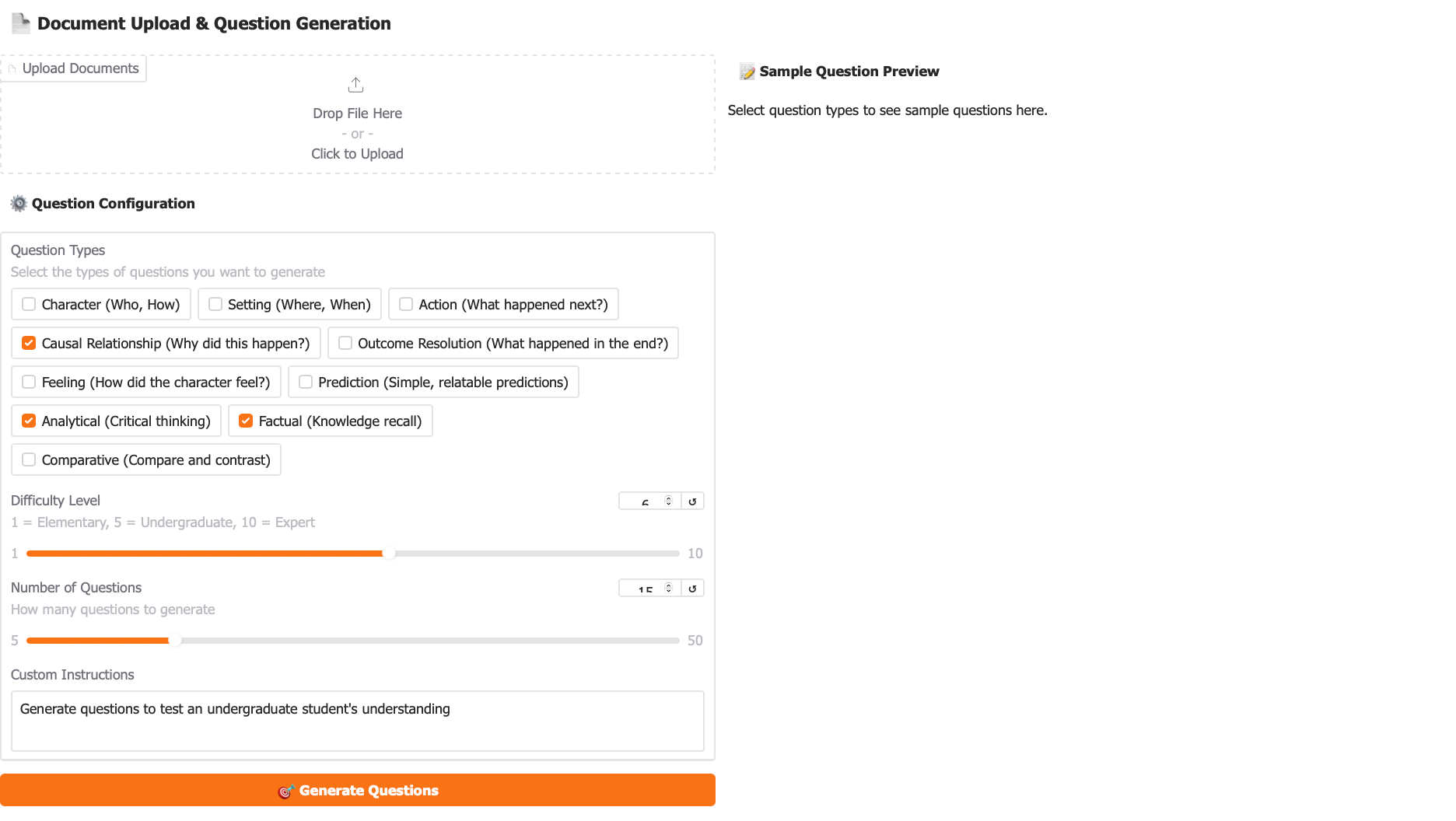}
  \caption{Interface of YourBench question generation panel.}
  \vspace{-0.5cm}
  \label{fig:interface}
\end{figure}

Our framework, named YourBench4Edu, is based on the pipeline of YourBench~\cite{shashidhar2025yourbencheasycustomevaluation}, which was originally designed for the generation of LLM question answering (QA) benchmarking. Our work adapts the question generation of YourBench to the early literacy education scenario. We aim to create question-and-answer pairs from stories to assess reading comprehension, to be used by a conversational agent. The resulting pairs of questions and answers can be used by educators for quick preparation of exam material. Furthermore, YourBench4Edu can also support YourBench in evaluating state-of-the-art LLMs for reading comprehension assessment.

% The interface of the YourBench question generation pipeline is shown in Figure~\ref{fig:interface}. Using the components of YourBench, including ingestion, summarization, single-shot question generation, multihop question generation, and lighteval, the framework is built to generate high-quality comprehension questions suitable for assessment of English reading comprehension for Kindergarten to second-grade students. The overall approach works as follows:

The interface of the YourBench question generation panel is shown in Figure~\ref{fig:interface}, where the question types, difficulty level, and number of questions can be easily customized. Using the components of YourBench, including ingestion, summarization, and question generation, our framework generates high-quality comprehension questions suitable for assessment of English reading comprehension. The overall approach works as follows:

First, the ingestion component takes the learning materials in various formats (e.g. PDF, HTML, MD). A language model then attempts to normalize raw materials into text in markdown format (i.e., MD).
Then, the summarization component chunks each ingested text document into pieces, prompts a language model to summarize each piece, and finally prompts the model to integrate the list of summaries into one well-structured text.

The next step before question generation is segmentation. Depending on the length of the text or the similarity of the sentences, this component creates chunks that are used to generate the questions.
Then, given the chunks, summaries, predefined question types (e.g., true-false, factual, analytical), and an instruction to gear the task to the specific proficiency, question generation prompts a language model to generate diverse single-shot or multi-hop questions of different question types at the targeted difficulty level. The difference between the generation of two genres is that: each single-shot question is based on only one chunk, while each multi-hop question is raised based on multiple chunks.

% Eventually, the lighteval component integrates the questions into a unified dataset.

\section{System Validation}
\label{sec:system_validation}

\begin{table*}[t]
    \centering
    \scriptsize
    \setlength{\tabcolsep}{4pt}
    
    \begin{tabular}{l | c c c c | c c c c} \toprule
        Method & \multicolumn{4}{c|}{MAP@N(Rouge-L F1)} & \multicolumn{4}{c}{MAP@N(BERTScore F1)} \\
        & Top 10 & Top 5 & Top 3 & Top 1 & Top 10 & Top 5 & Top 3 & Top 1 \\ 

        \midrule[1pt]
        \multicolumn{9}{c}{\textbf{Prior Work}} \\ 
        \midrule[1pt]

        FQAG \cite{yao-etal-2022-ais} 
            & 0.440/0.435 & 0.375/0.374 & 0.333/0.324 & 0.238/0.228 
            & 0.9077/0.9077 & 0.8990/0.8997 & 0.8929/0.8922 & 0.8768/0.8776 \\
        SQG \cite{dugan-etal-2022-feasibility} 
            & 0.460/0.455 & 0.392/0.388 & 0.344/0.337 & 0.234/0.242 
            & 0.9056/0.9062 & 0.8953/0.8955 & 0.8876/0.8878 & 0.8707/0.8723 \\
        DQAG \cite{eo-etal-2023-towards} 
            & 0.500/0.503 & 0.426/0.429 & 0.369/0.372 & 0.247/0.254 
            & 0.9156/0.9178 & 0.9046/0.9068 & 0.8956/0.8977 & 0.8752/0.8783 \\
        BART-large \cite{lewis-etal-2020-bart} 
            & 0.375/0.353 & 0.354/0.332 & 0.337/0.314 & 0.298/0.276 
            & 0.8911/0.8900 & 0.8878/0.8866 & 0.8851/0.8839 & 0.8794/0.8784 \\
        PFQS \cite{li-zhang-2024-planning} 
            & 0.569/\textbf{0.547} & 0.535/0.510 & 0.506/0.487 & 0.431/0.413 
            & \textbf{0.9198}/\textbf{0.9173} & \textbf{0.9144}/\textbf{0.9121} & \textbf{0.9099}/\textbf{0.9082} & 0.8988/\textbf{0.8965} \\

        \midrule[1pt]
        \multicolumn{9}{c}{\textbf{YourBench4Edu}} \\ 
        \midrule[1pt]

        Llama-3.3-70B-Instruct 
            & 0.515/0.497 & 0.508/0.489 & 0.500/0.481 & 0.477/0.459 
            & 0.9027/0.8996 & 0.9018/0.8987 & 0.9006/0.8977 & 0.8971/0.8943 \\
        Qwen3-235B-A22B 
            & 0.508/0.497 & 0.496/0.483 & 0.484/0.470 & 0.454/0.439 
            & 0.9031/0.9022 & 0.9015/0.8999 & 0.9000/0.8986 & 0.8953/0.8942 \\
        QwQ-32B 
            & \textbf{0.573}/0.532 & \textbf{0.553}/\textbf{0.516} & \textbf{0.540}/\textbf{0.502} & \textbf{0.504}/\textbf{0.466} 
            & 0.9107/0.9043 & 0.9083/0.9022 & 0.9064/0.9007 & \textbf{0.9011}/0.8960 \\

        \bottomrule
    \end{tabular}    
    \caption{A comparison between our method and prior work, using Rouge-L F1 and BERTScore F1 as the metrics.}
    \vspace{-0.5cm}

    \label{tab:comparison_top10}
\end{table*}

To validate the performance of our framework, we select the FairytaleQA validation set and test set~\cite{xu-etal-2022-fantastic} to carry out the assessment. The FairytaleQA is a QA corpus for narrative comprehension from kindergarten to eighth-grade student level, annotated by educational experts. To adapt our framework to the question generation evaluation scenario, where the system generates questions based on given ground-truth answers, we have two modifications to the vanilla pipeline. First, before the step of question generation, for each ground-truth question, we prompt the language model to determine the most relevant chunk given the ground-truth question-and-answer pair and the entire list of chunks obtained in the segmentation step. Second, during the question generation, the ground-truth answer is provided as an additional instruction in the question generation prompt.

We adopt MAP@N with Rouge-L F1 and MAP@N with BERTScore F1 as metrics, which were initially used by Eo et al.~\cite{eo-etal-2023-towards}. The results with different language models under the hood, including Llama-3.3-70B-Instruct, Qwen3-235B-A22B, and QwQ-32B, are compared with previous work, including FQAG~\cite{yao-etal-2022-ais}, SQG~\cite{dugan-etal-2022-feasibility}, DQAG~\cite{eo-etal-2023-towards}, BART-large~\cite{lewis-etal-2020-bart}, and PFQS~\cite{li-zhang-2024-planning}. Table~\ref{tab:comparison_top10} lists the results of the previous work and our experiments. Our framework shows a solid margin over the prior work under the MAP@N metric with Rouge-L F1. For MAP@N with BERTScore F1, our framework maintains a short gap with the best method. This suggests the state-of-the-art quality of the generated questions.

\section{Conclusion}
\label{sec:conclusion}
Evaluation of reading comprehension through content-related interactions is an essential component of the reading acquisition process. In this paper, we propose a novel approach for generating comprehension questions geared toward K-2 English learners. The proposed framework, YourBench4Edu, is based on YourBench and is able to ingest learning materials and generate a large diversity of question types at various difficulty levels. We also evaluate the performance of various language models in this framework using the FairytaleQA dataset as the source material and validate the quality of the generated questions. We believe that the proposed approach has the potential to become an important part of autonomous AI-driven English instructors.

% For camera-ready only, remember to comment it out
\section{Acknowledgements}
\label{sec:acknowledgement}
This paper is part of the project CELaRAI, which is supported by Institute of Education Sciences, U.S. Department of Education.

\bibliographystyle{IEEEtran}
\bibliography{mybib}

\end{document}